# Exploring New Directions in Iris Recognition

Nicolaie Popescu-Bodorin, *Member, IEEE*
Artificial Intelligence and Computational Logic Laboratory
Department of Mathematics and Computer Science
Spiru Haret University of Bucharest, Bucharest, Romania
http://fmi.spiruharet.ro/bodorin/

*Abstract*—A new approach in iris recognition based on *Circular Fuzzy Iris Segmentation* (CFIS) and *Gabor Analytic Iris Texture Binary Encoder* (GAITBE) is proposed and tested here. CFIS procedure is designed to guarantee that similar iris segments will be obtained for similar eye images, despite the fact that the degree of occlusion may vary from one image to another. Its result is a circular iris ring (concentric with the pupil) which approximates the actual iris. GAITBE proves better encoding of statistical independence between the iris codes extracted from different irides using *Hilbert Transform*. Irides from *University of Bath Iris Database* are binary encoded on two different lengths (768 / 192 bytes) and tested in both single-enrollment and multi-enrollment identification scenarios. All cases illustrate the capacity of the newly proposed methodology to narrow down the distribution of inter-class matching scores, and consequently, to guarantee a steeper descent of the *False Accept Rate*.

## I. Introduction

When a person's identity must be established or confirmed, iris recognition is the most reliable nonintrusive biometric technique to rely on. Since the early '90s when Wildes and Daugman pioneered this domain, it has become a popular subject and a constant preoccupation for scientific community. Still, the mathematical models underlying different iris recognition approaches proposed in the last two decades do not seem to become simpler, while the initial hypothesis that different irides match each other *by chance* seems to be forgotten. Here, we will prove that state-of-the-art iris recognition results can be obtained using a simpler mathematical model which also enables a better encoding of the statistical independence between iris codes of different irides, a steeper descent of the False Accept Rate and increased storage efficiency.

Because of the limited space available here, we would like to refer to Bowyer et al. for a survey of iris recognition [1], rather than including our own survey which would be neither shorter nor better.

### A. Outline

Basic computational routines used throughout this paper are presented in the second section. The third section of the paper proposes a new approach to iris segmentation. The basic idea of this approach is that since the pupil boundary is nearly circular, there must be a nearly circular pupil-concentric iris segment containing the muscles which control the pupil movements and play the most important role in iris recognition.

In other classical iris segmentation procedures, like those in Wildes's [27] and Daugman's approaches [2]-[10], iris segmentation means fitting circular contours by solving *three-dimensional* optimization problems to find a radius and two center coordinates via gradient ascent or by using edge detectors and Hough transform [14] or by iterating active contours [9],[10].

Here, each boundary is found mainly by defining and solving *one-dimensional* optimization problems and by using two basic computational procedures: k-Means Quantization [20],[22] and Run-Length Encoding [24].

The proposed Circular Fuzzy Iris Segmentation (CFIS) procedure guarantees that similar segmentation results will be obtained for similar eye images, despite the fact that the degree of occlusion may vary from one image to another. It consists of two steps: *pupil finding* and *limbic boundary approximation*. Both of them are fuzzy approaches based on k-Means and Run Length Encoding (RLE). The result of the procedure is a circular ring concentric with the pupil boundary. It is not the best approximation for one specific iris but it is a stable one, meaning that for a class of similar eye images representing the same eye, similar circular rings are obtained. When two similar images of the same eye are compared, each circular ring detected is pointing to the same physical support, possibly occluded by eyelids, eyelashes, specular and lighting reflections.

The fourth part of the paper describes the iris binary code extraction based on *Hilbert Transform*. The strong analytic signal associated with the chromatic iris sequence is used to recover phase information from iris texture in the angular direction. The reasons for avoiding the discovery of the radial iris features are also explained here. Gabor Analytic Iris Texture Binary Encoder (GAITBE) is introduced showing that accurate recognition of similar iris images can be achieved comparing the binary iris codes that the encoder will generate from the similar circular iris rings (segments) previously extracted through CFIS procedure.

A new approach to iris recognition based on CFIS and GAITBE is introduced in the fifth section. The proposed methodology is tested in both single-enrollment and multi-enrollment *identification* scenarios using two different lengths for iris codes. The results of these tests are presented and discussed in the sixth section of the paper.

## II. BASIC COMPUTATIONAL ROUTINES

**Run-Length Encoding** (RLE) is one of the simplest and most popular data compression algorithms [24]: for a given array, any subarray of redundant values is coded as a pair representing its histogram, as shown in the following example: if $V = [1,1,1,1,0,0,1,1,1,1,1,1,1,1]$ is the vector to be encoded, run-length encoding of V is:

$$rle(V) = [(1,4),(0,2),(1,8)].$$

**Run-Length Quantization for Binary Images** (RLQ) is defined here as a procedure for replacing all non-zero entries of a binary image with the corresponding run-length coefficients re-quantized in the unsigned 8-bit integer domain (uint8) by some custom quantization function, as illustrated in the following example:

$$V = [1,1,1,,0,0,1,1,1,1,1,1] \rightarrow [(1,3),(0,2),(1,6)] \rightarrow$$
$$[(3,3),(0,2),(6,6)] \rightarrow [(128,3),(0,2),(255,6)] \rightarrow$$
$$[128,128,128,0,0,255,255,255,255,255,255] = rlq(V),$$

where the third transform is the (re)-quantization function:

$$rqf(V) = \min(255, \max(1, \text{round}(255 \cdot V / \max(V)))).$$

RLQ procedure encodes a morphological property of the input image into a new signal (re-quantized image) by giving the same chromatic meaning to all of the white pixels sharing the same *run-length coefficient*.

**Fast k-Means Image Quantization** (FKMQ) is a variant of k-means algorithm designed for fast chromatic clustering in uint8 domain. It transforms the input image in an equipotential chromatic map [22] with k levels by replacing each chromatic value with the closest centroid. A suboptimal (incomplete) variant of FKMQ [20] can be easily derived by imposing termination in a small number of iterations while resetting the first centroid to the minimum available value (or to zero). In this way, the input image is forced to reveal its own range of darkness (numerical meaning of darkness according to the image histogram) and to return a handler to that area covered by lower chromatic values. This is particularly useful in detecting the pupil location.

## III. IRIS SEGMENTATION

### A. RLE-FKMQ Based Pupil Finder

Let us consider an eye image [26] like that in Fig.1.a. Its k-means equipotential chromatic map is revealed (Fig.1.b) by applying the FKMQ algorithm. The pupil cluster (PC) is then defined as the lowest level (cluster) on this map (Fig.1.c). It can be seen in Fig.1.c that the pupil cluster PC contains a good indication for the actual pupil perturbed by specular lights and eyelashes. The first problem to be solved at this stage is finding at least one pixel in PC which belongs for sure to the actual pupil. Defining an adaptive erosion procedure is the key for finding such a suitable pupil indicator: for each pixel within the pupil cluster, the structuring element of the erosion is adaptively determined by the morphological context in the neighborhood of that pixel, meaning that the vertical and horizontal run-length encoding is used to requantize (in uint8 domain) run-length coefficients corresponding to that pixel (i.e. the degree of membership of that pixel to a shorter or a longer continuous segment lying in the pupil cluster).

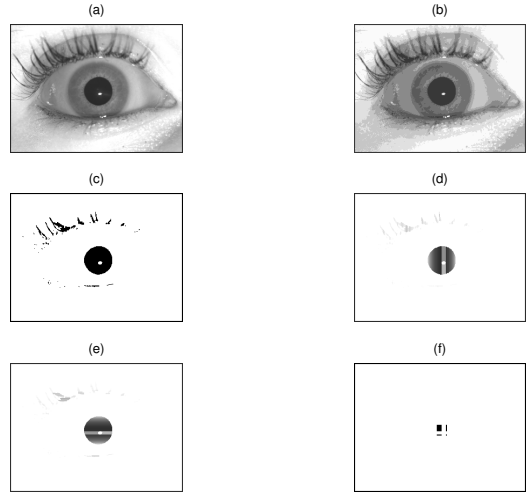

Figure 1. Original eye image (a) and its 8-means quantization (b); The pupil cluster PC (c); Vertical run-length quantization of the pupil cluster (d); Horizontal run-length quantization of the pupil cluster (e); The pupil indicator PI (f). Images from (c) to (f) are presented in complement.

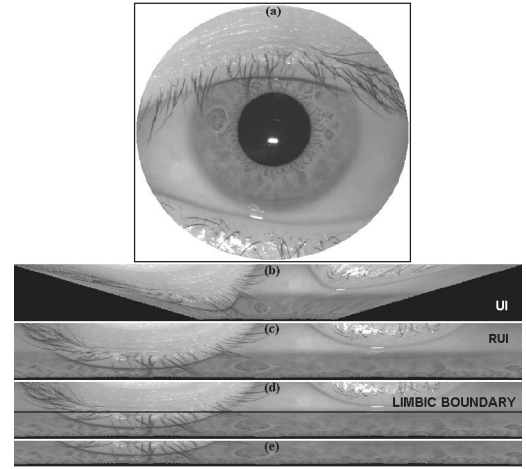

Figure 2. Iris segmentation stages of the proposed approach.

Consider matrices RLV (Fig.1.d) and RLH (Fig.1.e) as being the vertical and horizontal run-length quantization of the pupil cluster PC, respectively. As they are 8-bit gray images, the FKMQ algorithm will be applied to retrieve the darker area in both of them. The results are then combined to form the pupil indicator PI (Fig.1.f).

Any pixel within the pupil indicator can now be used as a starting point for a flood-fill operation. The most accurate pupil segment available in the pupil cluster is identified this way. Further, the specular lights are filled using run-length encoding once again. The result is then fitted into a rectangle and approximated by an ellipse or a circle (see pupil boundary in Fig.6).

The type and extent of the specular lights depend on the illumination scheme used during the image acquisition process. The eye images used here [26] contain only single-spot specular lights which are easy to fill using the following rule: until the pupil does not change anymore, those (groups of) pixels whose neighbours (on a horizontal / vertical line) belong to the pupil are to be filled. Other databases could require a different procedure for filling specular lights (CASIA-V3-Interval, for example).

Summarizing the operations described above, the proposed RLE-FKMQ Based Pupil Finder procedure can be stated as follows:

**RLE-FKMQ Based Pupil Finder** (N. Popescu-Bodorin)**:**
INPUT: the eye image IM;
1. Extract the pupil cluster:
   PC = fkmq(IM,16);
   PC = (PC == min(PC));
2. Compute horizontal and vertical RLQ of PC:
   RLV(:,j) = vrleq(PC);
   RLH(j,:) = hrleq(PC);
3. Compute the pupil indicator PI:
   [k, PI] = getpi(RLH, RLV);
   PI = find(PI == 1);
   PI = PI(1);
4. Extract available pupil segment through a flood-fill operation:
   P = imfill(PC, PI);
5. Fill the specular lights:
   P = rlefillsl(P);
6. Approximate the pupil by an ellipse;
OUTPUT: The ellipse approximating the pupil;
END.

*B. Circular Fuzzy Iris Segmentation*

RLE-FKMQ Based Pupil Finder procedure guarantees accurate pupil localization and enables us to unwrap the eye image (Fig.2.a) in polar coordinates (Fig.2.b) and also to practice the localization of the limbic boundary in the rectangular unwrapped eye image (Fig.2.c), obtaining an iris segment as in Fig.2.e.

**Circular Fuzzy Iris Segmentation** (N. Popescu-Bodorin):
INPUT: the eye image IM;
1. Apply RLE-FKMQ Based Pupil Finder procedure to find pupil radius and pupil center;
2. Unwrap the eye image in polar coordinates (pixel-to-pixel polar transcoding; see the unwrapped image *UI* in Fig.2.b);
3. Stretch the unwrapped eye image *UI* to a rectangle (*RUI* - Fig.2.c);
4. Compute three column vectors: *A*, *B*, *C*, where *A* and *B* contain the means of the lines within *UI* and *RUI*, respectively. *C* is the mean of the lines within the [*A B*] matrix;
5. Compute *P*, *Q*, *R* as being 3-means quantizations of *A*, *B*, *C* (Fig.3);
6. For each line of the unwrapped eye image count the votes given by *P*, *Q* and *R*. All the lines receiving at least two positive votes are assumed to belong to the actual iris segment.
7. Find limbic boundary and extract the iris segment (Fig.3, Fig.2.d, Fig.2.e);
OUTPUT: pupil center, pupil radius, index of the line representing limbic boundary and the final iris segment;
END.

The pupil indicator is found as a preimage corresponding to the maximum value of a fuzzy membership assignment describing the actual pupil as a subset of the pupil cluster: for each pixel within the pupil cluster, directional run-length coefficients encode the degree of membership of that pixel to the actual pupil. The argument is that being (or containing) the most circular solid object within the pupil cluster, the actual pupil is the most resilient set to erosion [18] that can be found in the pupil cluster.

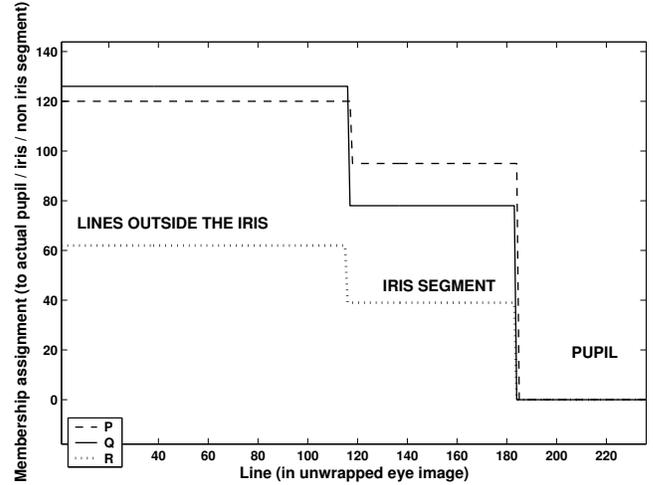

Figure 3. Iris segmentation procedure: line assignment (step 5).

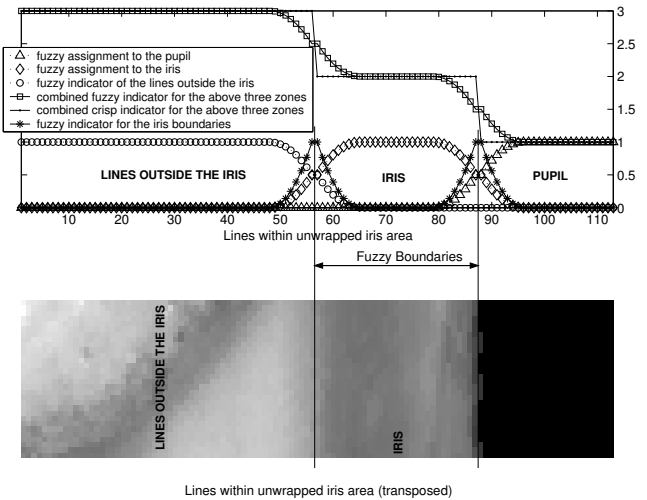

Figure 4. Fuzzy iris segment and fuzzy iris boundaries.

Also, the computation of the pupil indicator depends on a single parameter: a threshold for the horizontal and vertical run-length coefficients computed for the pupil cluster, a threshold above which the membership of a pixel to the actual pupil is guaranteed.

For these two reasons, the proposed pupil finder procedure is a fuzzy approach that solves a one-dimensional optimization problem.

On the other hand, Fig.4 shows what happens to the vectors *A*, *B* and *C* at steps 4-5 of the CFIS procedure: behind the combined crisp indicator function (crisp membership assignment) of any 3-means quantization (Fig.3), there are fuzzy membership assignment functions defined from the set of lines within the rectangular unwrapped iris area (*RUI*) to the pupil, the iris, the area outside the iris and even to the iris boundaries. Hence, there is no doubt that the area delimited between the fuzzy iris boundaries (Fig.4) is the fuzzy iris band. Its preimage through the polar mapping is a circular fuzzy iris ring. Three fuzzy iris bands are determined using vectors *A*, *B*, *C*. The final result is computed evaluating the odds that the lines within the unwrapped iris area belong to the actual iris segment. This is done in step 6 of the CFIS procedure by counting the votes received for each line within the unwrapped iris area as a member of a fuzzy iris band. Also, there is no doubt that the limbic boundary is determined by searching for a line number. For these reasons, the proposed CFIS procedure is a fuzzy approach that solves a one-dimensional optimization problem.

In fuzzy set theory [23], a membership function that only takes binary values is called a crisp indicator function. We extend the meaning of this definition by making the following considerations: a crisp indicator is, in fact, the ordinary indicator function of an ordinary sub-set within a set:

$$I_A : X \to \{0,1\}; \forall a \in X, I_A(a) = logical(a \in A).$$

For any sub-set *A* of *X*, $X = A \cup \overline{A}$, (where $\overline{A}$ denotes the complement of *A* in *X*), hence we may consider that the crisp indicator of *A* is nothing more than an encoding (in two symbols) of a disjoint cover of *X* containing two sets: *A* and its complement (regardless the nature or the values of those two symbols and the nature of the sets *A* and *X*). Consequently, it is naturally to define *combined crisp indicator of a disjoint reunion*

$$X = \bigcup_{j=1}^{n} A_j,$$

as being the sum:

$$CCI_X = \sum_{j=1}^{n} j \cdot I_{A_j}, \qquad (1)$$

or more generally, as follows:

$$\forall k \in \overline{1,n}, \forall a \in A_k, CCI_X(a) = S(\sum_{j=1}^{n} j \cdot I_{A_j}(a)) = s_k, \quad (2)$$

where $S = \{s_j\}_{j \in \overline{1,n}}$ is a sequence of distinct symbols.

It means that a combined crisp indicator of a disjoint reunion is unique up to a bijective correspondence between the sequences of symbols that are used to encode the memberships to each set within the reunion. Hence, if *X* is restricted to **R**, the combined crisp indicator of a disjoint cover of *X* is exactly the equivalence class of all step functions that can be defined using the sets of that cover. If *X* is a discrete signal, then we resort to discrete step functions. Consequently, any discrete step function is equivalent (in the above defined sense) to a combined crisp indicator (1), and in particular, any k-means quantization of a one-dimensional signal (like *P*,*Q* and *R* in Fig.3) is equivalent to a combined crisp indicator (like that in Fig.4). Therefore, it doesn't really matter what symbols (or values) are used to encode the crisp indicator function. In Fig.3, combined crisp indicators are encoded using chromatic values because, otherwise, it could be difficult to distinguish between them.

## IV. GABOR ANALITIC IRIS TEXTURE BINARY ENCODER

### A. Hilbert Transform and the strong analytic signal

The strong analytic signal was introduced by Gabor [11] for extracting phase information content from a finite, discrete signal given in time domain i.e. for recovering both the carrier wave and the message modulated on it from the given signal.

For a continuous time-domain signal *f(t)*, its Hilbert Transform is defined as follows:

$$H(f(t)) = \frac{1}{\pi} P \int_{-\infty}^{\infty} \frac{f(\tau)}{t-\tau} d\tau, \qquad (3)$$

when the integral exists (in terms of Cauchy principal value).

A strong analytic signal is the complex continuous time-domain signal *f(t)* having the following property:

$$H(f(t)) = \frac{1}{\pi} P \int_{-\infty}^{\infty} \frac{f(\tau)}{t-\tau} d\tau = -j \cdot f(t).$$

If the strong analytic signal *f(t)* is split into real and imaginary parts:

$$f(t) = g(t) + j \cdot h(t),$$

then:

$$H(f(t)) = \frac{1}{\pi} P \int_{-\infty}^{\infty} \frac{g(\tau)}{t-\tau} d\tau + j \cdot \frac{1}{\pi} P \int_{-\infty}^{\infty} \frac{h(\tau)}{t-\tau} d\tau = h(t) - j \cdot g(t),$$

and therefore:

$$H(f(t)) + j \cdot H(f(t)) = h(t) - j \cdot g(t)$$

Hence,

$$H(f(t)) = h(t), \; H(f(t)) = -g(t),$$

or in other words:

$$H(Re(f(t))) = Im(f(t)),$$

Consequently, any signal:

$$y = x + j \cdot H(x), \qquad (4)$$

where *H* denotes the Hilbert Transform (3), is a strong analytic signal (Gabor analytic signal associated with *x*).

The *analytic image* [12] is the 2-dimensional version of the strong analytic signal (having the same form as mentioned in (4) where x is an image instead) and can be used for iris recognition as in [25]. Inspired and motivated by this work, the currently proposed Gabor Analytic Iris Texture Binary Encoder is a simpler and more robust approach to iris binary code extraction based on the discovery of phase information available in the iris texture.

The main reason for working with one-dimensional strong analytic signal instead of using the analytic image is that the critical information which decides on the similarity or non-similarity between two iris rings is mainly stored as chromatic variation in the angular direction. On the other hand, accurate iris movement equations should be available in order to trace and to match chromatic variations along the radial direction. Before knowing such motion laws, the chromatic variation along the radial direction will be without any doubt an important source of disagreement between those circular iris rings (segments) representing the same iris in different pupil dilatations. The essence of such a disagreement is that it is not "reconcilable" through an elastic deformation. As a practical example one can consider an iris image in which pupil dilatation is sufficiently strong to cause the iris area closest to the pupil to "disappear" or to change dramatically.

### B. Recovering phase content from Gabor analytic signal

In the context of iris recognition, the most important property of the Hilbert Transform is that it preserves the signal energy. Hence f and H(f) have the same energy but also $\frac{d}{dt}H(f), H(\frac{d}{dt}f), \frac{d}{dt}f$ share the same energy. When $f$ is assumed to be a line within the unwrapped iris, the meaning of the above fact is that the iris features in the angular direction are encoded with the same fidelity both in $f$ and $H(f)$.

Now, let us consider the real time-domain signal $f(t)$, its Hilbert Transform $H(f)$ and associated Gabor analytic signal:
$$z(t) = f(t) + j \cdot H(f(t))$$
expressed in polar form:
$$z(t) = A(t)e^{j\Phi(t)},$$
where:
$$A(t) = \sqrt{(f(t))^2 + (H(f(t)))^2}$$
is the instant amplitude and:
$$\Phi(t) = arctan(\frac{H(f(t))}{f(t)})$$
is the instant phase, further used to generate iris binary code.

### C. Encoding the unwrapped iris

The unwrapped iris (Fig.2.e) is obtained by applying the CFIS procedure. Each line within the unwrapped iris is associated with a corresponding Gabor analytic signal (4).

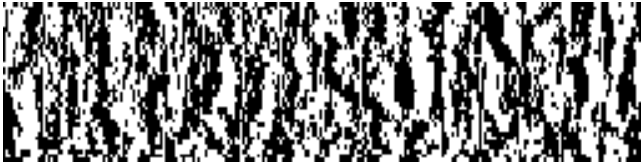

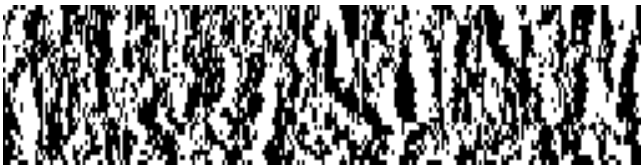

Figure 5. Two similar iris codes obtained for similar iris images ([26], 0001-L-0001.j2c, 0001-L-0003.j2c) using Gabor Analytic Iris Texture Binary Encoder

As a practical example, the iris codes extracted from two images ([26], 0001-L-0001.j2c, 0001-L-0003.j2c) are presented in Fig.5. Hamming similarity measure for the two iris codes in Fig.5 is 0.7346. The binary iris code is generated as follows:

**GAITBE procedure** (N. Popescu-Bodorin):
INPUT: unwrapped iris image *IM* (Fig.2.e), window dimension *s*, desired dimension $d = [d_1, d_2]$ for iris codes;
1. Compute *I* as being the resized version of unwrapped iris segment to desired dimension *d*;
2. For each line of *I* compute the complex matrix *AS* as being the strong analytic representation of *I* using window size *s*;
3. For *AS*, compute the instant phase matrix *IP*;
4. Compute the binary iris code: $IC = logical(IP \geq 0)$;
5. Compute the binary iris mask *M* corresponding to the various iris occlusions (specular lights, diffuse reflections, eyelashes, eyelids, etc), if any;

OUTPUT: The binary iris code *IC* and its binary mask *M*;
END.

### V. PROPOSED IRIS RECOGNITION SYSTEM ARHITECTURE

It is required for our iris recognition system to have a minimal list of operating modes (acquisition, calibration, enrollment, verification, identification), each of them dealing with a specific task: acquire the image, extract the iris, encode the iris texture, store iris template, matching between iris codes. For this reason, all major functionalities are implemented here as modules. The module collection mirrors the evolution stages from an experimental recognition system to a real-world application. It is also required that the proposed system should be a supervised intelligent agent, able to explore, identify and test new directions, to mine available knowledge and data. A part of this architecture, covering five functionalities, will be discussed here:

**NPB Iris Recognition Generic Experimental Model** (N. Popescu-Bodorin):

**Segmentation Module:** CFIS;
**Encoding Module:** GAITBE;

**Calibration/Training Module**:
INPUT: a collection of eye images divided in classes (each class corresponds to one specific eye of one specific person), desired dimension $d = [d_1, d_2]$ of iris codes, window dimension *s*, other control parameters;
1. For each image within the given collection perform the following three operations:
   - Run Segmentation Module to compute pupil center, pupil and iris radii (a circular iris ring approximating the actual iris segment), unwrapped iris segment;
   - Run Encoding Module to compute the iris code *C* (binary quantization of the phase information available in the unwrapped iris segment) and the occlusion mask *M*;
   - Store the iris code, corresponding occlusion mask and other data, if required;
2. Compute the experimental and theoretical distributions of intra-class / inter-class similarity scores and infer a recognition threshold;
3. Use the above information to compute experimentally determined FAR, FRR, EER, (False Accept/Reject Rate, Equal Error Rate), theoretical Odds of False Accept/Reject (OFA, OFR), other coefficients reflecting different evaluation criteria: Daugman's decidability index [4], Fisher's ratio [27], storage efficiency, or the value of FRR for a given FAR of 0.001 ([16],[19],[28]);

ATEXIT: Launch Analyzer Module posting all data handlers to it;

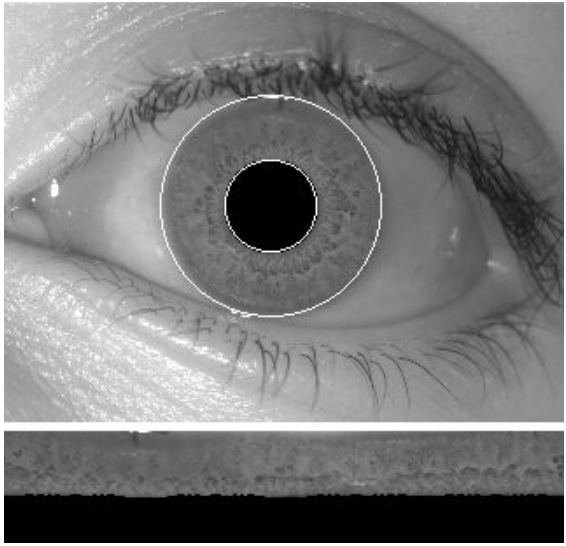

Figure 6. Circular Fuzzy Iris Segmentation Demo Program.

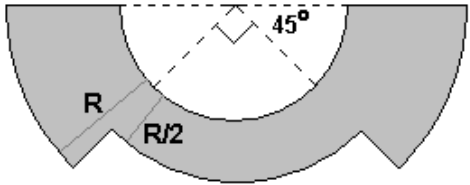

Figure 7. Butterfly Iris Segment.

**Analyzer/Advisor Module:**
INPUT: Current data, History data;
OUTPUT: Display all coefficients used to evaluate system performance. Suggest a recognition threshold or an adjustment of control parameters. Store the system parameters history for later use. Recommend the next task.
REQUEST: User choice for next action: accept/modify automatically determined recognition threshold and/or suggested control parameters adjustment; accept/change the next task;
ATEXIT: Launch next task and transfer required data handlers to the module designed to perform that task;

**Iris Enrollment Module:**
INPUT: the number of left/right eye images (or corresponding iris codes) to be enrolled under the same identity as biometric templates, eye image source / iris code datasource, a rule for choosing between available iris codes (random, maximize inter-class similarity scores distribution, minimize intra-class similarity scores distribution, etc.), other control parameters;
For each image given to be enrolled, perform the following operations: run Segmentation Module; run Encoding Module; store the iris code, corresponding occlusion mask and other data, if required; maintain the list of enrolled / unenrolled images;
For each unenrolled image compute the distance to all enrolled classes;
Compute the experimental and theoretical distributions of intra-class / inter-class similarity scores and infer a recognition threshold;
Compute FAR, FRR, EER, OFA, OFR, other coefficients reflecting evaluation criteria;
ATEXIT: Launch Analyzer Module posting all data handlers to it;

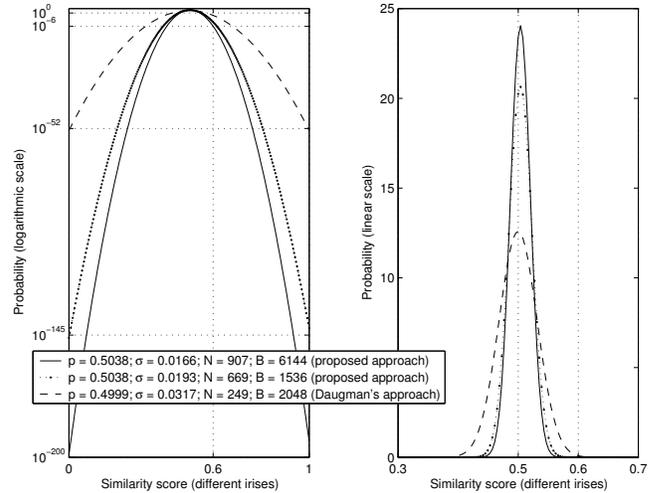

Figure 8. Comparing the distributions of inter-class matching scores.

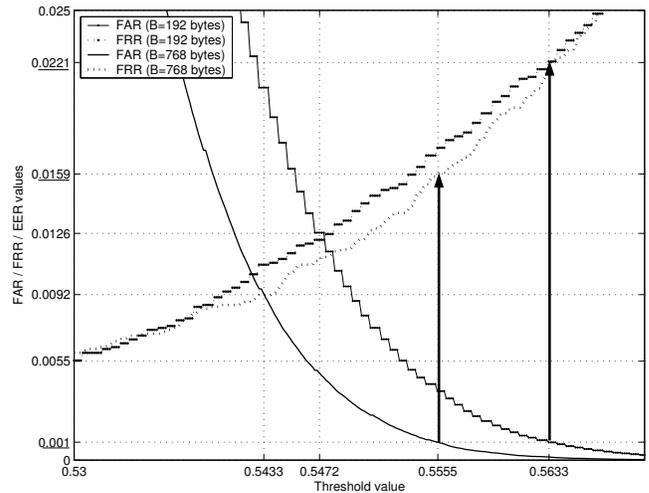

Figure 9. FAR-FRR variation for the tests (T1), (T2).

## VI. IMPLEMENTATION DETAILS. ANALYSIS OF EXPERIMENTAL DATA

The software modules described above are currently implemented using Matlab and calibrated for the database [26]. CFIS demo program (Fig.6) is available for download [20].

Using 4258 different iris images, Daugman [8] shows that in his model the distribution of Hamming distances between different irides matches a binomial distribution with $p = 0.5$ and $N = 249$ degrees-of-freedom, or a normal distribution around the mean $p = 0.499$ with standard deviation $\sigma = 0.0317$. $N$ and $\sigma$ ($N = p(1-p)/\sigma^2$) express the amount of difference between iris codes of different irides as a result of correlated Bernoulli trials with 249 tosses of a fair coin. He also shows that if all 2048 bits in an iris code were independent then the standard deviation would be $\sigma = 0.011$.

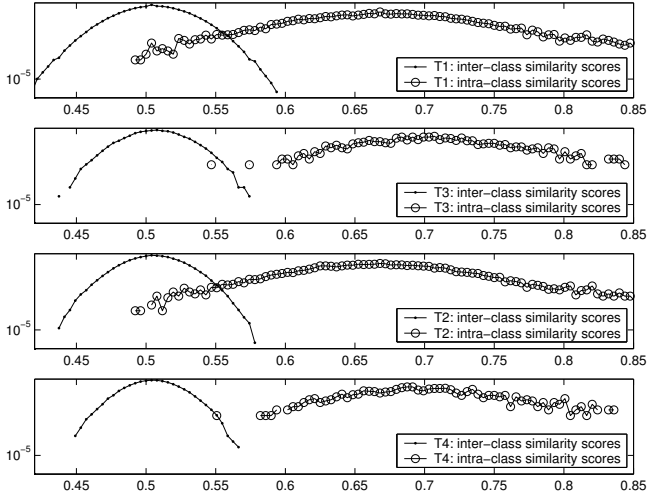

Figure 10. Score distributions for all tests (log scale).

TABLE I. STATISTICS OF EXPERIMENTAL DATA.

| Test | (T1) | (T2) | (T3) | (T4) |
|---|---|---|---|---|
| Iris code length: | 192 B | 768 B | 192 B | 768 B |
| Window size: | 8 | 16 | 8 | 16 |
| Enrolled templates | 1 | 1 | 5 | 5 |
| **Inter-class scores:** | | | | |
| Mean: | 0.5038 | 0.5038 | 0.5049 | 0.5031 |
| Median: | 0.5039 | 0.5037 | 0.5045 | 0.5030 |
| Standard deviation: | 0.0193 | 0.0166 | 0.0169 | 0.0152 |
| Degrees-of-freedom: | 669 | 908 | 874 | 1089 |
| Skewness: | -0.008 | 0.0024 | -0.007 | 0.0266 |
| Kurtosis: | 0.0366 | 0.0488 | 0.0393 | 0.0465 |
| **Intra-class scores:** | | | | |
| Mean: | 0.6718 | 0.6678 | 0.7039 | 0.6993 |
| Median: | 0.6706 | 0.6659 | 0.7019 | 0.6962 |
| Standard deviation: | 0.0550 | 0.0533 | 0.0464 | 0.0467 |
| Degrees-of-freedom: | 73 | 78 | 96 | 96 |
| Skewness: | 0.3528 | 0.4547 | 0.2203 | 0.2855 |
| Kurtosis: | 1.0995 | 1.3842 | 0.1288 | 0.1670 |
| **Functioning regimes:** | | | | |
| **1.** At FAR=0.001: | | | | |
| Threshold value: | 0.5633 | 0.5555 | 0.5570 | 0.5503 |
| FRR: | 0.0221 | 0.0159 | 0.0013 | 0.0013 |
| **2.** At FRR=0.01: | | | | |
| FAR: | 0.0243 | 0.0072 | 0 | 0 |
| OFA: | 0.0232 | 0.0068 | 2.1E-9 | 2E-10 |
| OFR: | 0.0092 | 0.0105 | 0.0160 | 0.0155 |
| Treshold: | .54231 | .54475 | .60429 | .59847 |
| **3.** At threshold = 0.59: | | | | |
| FRR: | 0.0564 | 0.0587 | 0.0027 | 0.0054 |
| OFR: | 0.0682 | 0.0721 | 0.0071 | 0.0093 |
| FAR: | 2.1E-6 | 0 | 0 | 0 |
| OFA: | 4.1E-6 | 1.0E-7 | 2.4E-7 | 4.9E-9 |
| **4.** At threshold = 0.60: | | | | |
| FRR: | 0.0806 | 0.0853 | 0.0081 | 0.0107 |
| OFR: | 0.0955 | 0.1016 | 0.0126 | 0.0163 |
| FAR: | 0 | 0 | 0 | 0 |
| OFA: | 3.2E-7 | 3.3E-9 | 9.3E-9 | 8E-11 |
| **Evaluation criteria:** | | | | |
| Decidability: | 4.0794 | 4.1555 | 5.6495 | 5.6943 |
| Fisher's ratio: | 8.3208 | 8.6340 | 15.958 | 16.212 |
| EER: | 0.0159 | 0.0092 | 1.4E-3 | 1.3E-3 |
| Storage efficiency: | 0.4355 | 0.1478 | 0.5690 | 0.1772 |

Here, 1000 different images from Bath University Iris Database [26] are used to perform the following tests:

(**T1**): The Calibration module is tested in identification scenario (all-to-all images) using the following list of parameters: iris code length of 192 bytes, window size of 8 pixels, Hamming similarity measure (number of pixels that agree / number of compared pixels).

(**T3**): The Enrollment Module is tested in a multi-enrollment identification scenario with iris code length of 192 bytes and window size of 8 pixels. From the 20 images available for each eye, 5 images are used to generate 5 enrolled templates (an identity). The remaining 15 images (for each eye) are tested for identification: all unenrolled images are compared with all enrolled identities using the *mean-deviation similarity score* defined below.

(**T2**) & (**T4**): The tests are derived from (T1) & (T3), respectively, by changing iris code length to 768 bytes and window size to 16 pixels.

Suppose that the value $s$ of the standard deviation is known for the imposter score distribution in the single-enrollment scenario and let $C$ be the current input iris code which must be compared to a (finite) set of templates $E = \{E_k\}_{k \in \overline{1,n}}$ enrolled under an arbitrary identity $E$, and let $S$ be the set of Hamming similarities between $C$ and each of the enrolled templates: $S = \{HS(C, E_k)\}_{k \in \overline{1,n}}$. Then the *mean-deviation similarity score* between the input template $C$ and an arbitrary identity $E$ is heuristically defined here as:

$$MDS(C, E) = \text{mean}(S) + \text{std}(S) - s/2.$$

After a visual examination of Bath, ICE-2005 and CASIA-v.1-3 iris databases, the 'butterfly' iris segment (Fig.7) was found to be more often unoccluded. In order to ensure further compatibility, all tests here use this type of iris segment. Besides, for all images within the tested database [26], the occlusions over the butterfly iris segment have proven to be negligible.

The results of the tests (T1)-(T4) are presented in Table I and in Fig.8-10.

For a given threshold, the False Accept (Reject) Rate is experimentally determined here as the ratio between the number of imposter (genuine) scores exceeding (not exceeding) the threshold and the total number of imposter (genuine) scores.

The Odds of False Accept (Reject) are given by the cumulative of the theoretical imposter (genuine) distribution above (below) the given threshold. They approximate the definite integrals:

$$\text{OFA}(t) = \int_t^1 I_{\text{pdf}} d\tau, \quad \text{OFR}(t) = \int_0^t G_{\text{pdf}} d\tau,$$

where $I_{\text{pdf}}$ and $G_{\text{pdf}}$ are theoretical probability density functions of the imposter and genuine distributions, and $t$ is the threshold.

For tests (T1,T2), the variations of the False Accept/Reject Rates are presented in Fig.9, where FRR values at a FAR of 0.001 (i.e. 0.0159 and 0.0221) and EER values (0.0092, 0.0159) are also marked. Other reference values of this kind can be found in: Fig.3 in [19] (NIST Iris Challenge Evaluation for the following algorithms: CAM-2-Cambridge, IrTch-2-IriTech, SI-2-Sagem-Iridian), Table 1 in [28], Fig.5 in [16], Fig.5 in [14], Fig.9 in [13].

*Storage efficiency* is a new evaluation criterion proposed here. It is defined as the ratio between the number of degrees-of-freedom of inter-class score distribution and the length of the iris code.

By comparing tests (T1,T3) and (T2,T4), it can be seen that when identities are defined through a collection of enrolled templates (T3, T4), each identity includes some degree of variability (iris rotation, different pupil dilatations, different focalization and illumination). Consequently, the intra-class distribution becomes narrower and so does the inter-class distribution (Fig.10).

## VII. Conclusion and Future Work

This article has presented a novel approach to iris recognition. The proposed CFIS and GAITBE procedures both proved their capacity to narrow down the distribution of inter-class matching scores, to guarantee a steeper descent of the False Accept Rate and to achieve better encoding of statistical independence between the iris codes of different irides. Also, this is the first time that pupil finding and limbic boundary approximation are treated as one-dimensional optimization problems. More details and new results concerning this novel iris recognition approach will be available as soon as possible in the future work.

## Acknowledgment

The author wishes to thank Professor Luminita State (University of Pitesti, RO) for the comments, criticism, and constant moral and scientific support during the last two years. The author would also like to thank Professor Donald Monro (University of Bath, UK) for granting the access to the Bath University Iris Database. Without their support this iris recognition study could not have been possible. Thanks are also due to three anonymous reviewers for their very insightful comments.